\documentclass{article}

\usepackage{microtype}
\usepackage{graphicx}
\usepackage{subcaption}
\usepackage{booktabs} 

\usepackage{hyperref}

\usepackage[preprint]{icml2026}

\usepackage{amsmath}
\usepackage{amssymb}
\usepackage{mathtools}
\usepackage{amsthm}

\usepackage[capitalize,noabbrev]{cleveref}

\theoremstyle{plain}

\theoremstyle{definition}

\theoremstyle{remark}

\usepackage[textsize=tiny]{todonotes}

\icmltitlerunning{STEER: Inference-Time Risk Control via Constrained Quality-Diversity Search}

\begin{document}

\twocolumn[
  \icmltitle{STEER: Inference-Time Risk Control via Constrained Quality-Diversity Search}



  \icmlsetsymbol{equal}{*}

  \begin{icmlauthorlist}
    \icmlauthor{Eric Yang}{Verily Life Sciences}
    \icmlauthor{Jong Ha Lee}{Verily Life Sciences}
    \icmlauthor{Jonathan Amar}{Verily Life Sciences}
    \icmlauthor{Elissa Ye}{Verily Life Sciences}
    \icmlauthor{Yugang Jia}{Verily Life Sciences}
  \end{icmlauthorlist}

  \icmlaffiliation{Verily Life Sciences}{Verily Life Sciences, Dallas, TX, USA}

  \icmlcorrespondingauthor{Eric Yang}{eryang@verily.com}

  \icmlkeywords{Large Language Models, Clinical Triage, Quality-Diversity Search, Steerability, Alignment, Inference-Time Control, Ordinal Decision Making, Evolutionary Algorithms, ICML}

  \vskip 0.3in
]



\printAffiliationsAndNotice{}  

\begin{abstract}

Large Language Models (LLMs) trained for average correctness often exhibit mode collapse, producing narrow decision behaviors on tasks where multiple responses may be reasonable. This limitation is particularly problematic in ordinal decision settings such as clinical triage, where standard alignment removes the ability to trade off specificity and sensitivity (the ROC operating point) based on contextual constraints. We propose STEER (Steerable Tuning via Evolutionary Ensemble Refinement), a training-free framework that reintroduces this tunable control. STEER constructs a population of natural-language personas through an offline, constrained quality–diversity search that promotes behavioral coverage while enforcing minimum safety, reasoning, and stability thresholds. At inference time, STEER exposes a single, interpretable control parameter that maps a user-specified risk percentile to a selected persona, yielding a monotonic adjustment of decision conservativeness. On two clinical triage benchmarks, STEER achieves broader behavioral coverage compared to temperature-based sampling and static persona ensembles. Compared to a representative post-training method, STEER maintains substantially higher accuracy on unambiguous urgent cases while providing comparable control over ambiguous decisions. These results demonstrate STEER as a safety-preserving paradigm for risk control, capable of steering behavior without compromising domain competence.
\end{abstract}

\section{Introduction}

\begin{figure*}[ht] 
    \centering
    \includegraphics[width=0.9\textwidth]{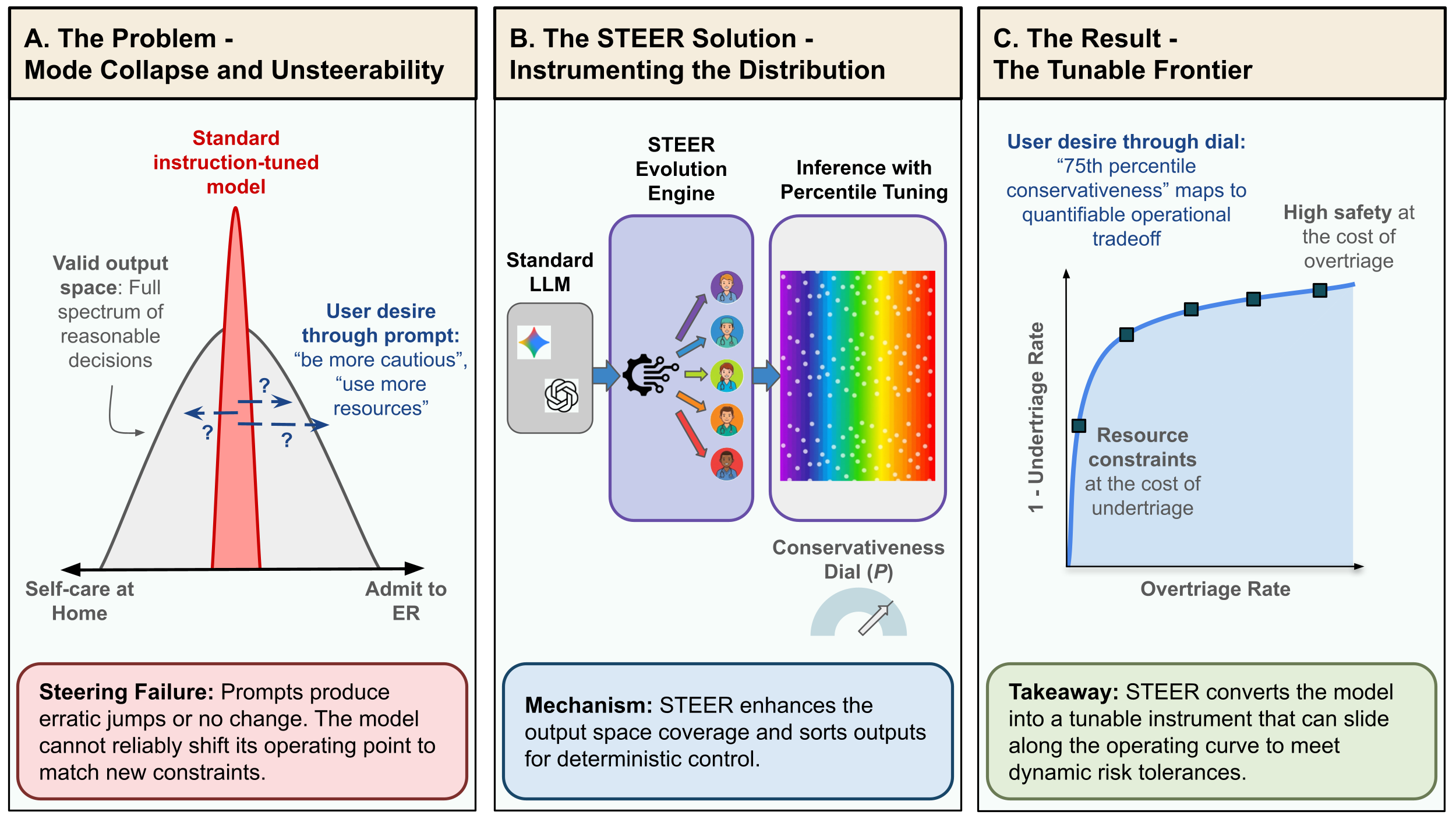}
    \caption{\textbf{STEER Reintroduces the Risk Operating Point. (A) Problem:} Standard models exhibit mode collapse, shrinking the distribution of valid disagreement (gray curve). They are also unsteerable, reacting erratically to prompts (blue arrows). \textbf{(B) The STEER Framework:} STEER enhances this distribution via a constrained quality-diversity evolutionary search. This creates a deterministic conservativeness dial ($P$). \textbf{(C) Context-Aware Control:} The system generates a steerable ordinal performance curve (analogous to an ROC), enabling users to select a precise operating point ($P$) that satisfies specific sensitivity/specificity requirements.}
    \label{fig:Overview}
\end{figure*}

Large Language Models (LLMs) are rapidly being deployed as decision-support tools in high-stakes domains such as clinical triage, financial planning, and legal judgments \cite{tyler2024use, Lee2025-xu, Rashidian2025-ml, vukovic2025ai, kolkman2024justitia}. In these settings, models are decision partners that must align with the user-specific operational constraints. However, a critical limitation that prevents their safe, widespread deployment is their inability to consistently steer model behavior along a risk spectrum. In traditional machine learning, optimality is managed by selecting an operating point on the Receiver Operating Characteristic (ROC) curve, allowing operators to explicitly trade off specificity and sensitivity based on contextual needs \cite{Stanfield2015-vz}.  For instance, a resource-rich hospital may prioritize minimizing undertriage, whereas a capacity-strained facility requires stricter criteria \cite{justice2025ems}. Consequently, the utility of an AI decision support system depends not only on its accuracy but on its tunable risk tolerance, which is the ability to reliably adjust its operating point along a continuum of valid judgments.

Current paradigms for aligning LLMs, such as Reinforcement Learning from Human Feedback (RLHF), are fundamentally at odds with this requirement. By training models to maximize average correctness, these techniques inadvertently induce mode collapse in ambiguous scenarios \cite{Meng2024-bv, Kirk2023-er, Hamilton2024-ja}. The optimization process sharpens the predictive distribution around a single response, effectively collapsing the rich spectrum of plausible expert interpretations into a low-entropy point estimate. This locks models into a static, one-size-fits-all risk posture that fails to represent the tail opinions crucial for safety-critical deliberation. Furthermore, standard prompting offers no deterministic mechanism to navigate latent diversity. While automated prompt optimization can discover a static configuration, it fails to provide the continuous control required for runtime adjustment.

In this work, we introduce \textbf{Steerable Tuning via Evolutionary Ensemble Refinement (STEER)}, a framework that unlocks the ability to select a precise operating point without retraining (Figure \ref{fig:Overview}). We formulate steerability as a distributional search problem. Unlike standard sampling methods that uncontrollably maximize entropy which often leads to hallucinations, STEER employs a constrained quality-diversity search. This evolutionary process explicitly optimizes for behavioral coverage (filling gaps in the risk spectrum) while strictly enforcing minimum thresholds for safety, reasoning-coherence, and variance regularization. At inference time, we introduce a mechanism that maps the ensemble's ordinal outputs to a single scalar parameter, effectively enabling the risk dial for operational control. Our main contributions are summarized as follows:
\begin{itemize}
  \item \textbf{Quality-Diversity for Output Space Coverage:} We introduce a constrained evolutionary search that systematically builds an ensemble to fill voids in the model's estimated risk distribution.
  \item \textbf{Deterministic Inference Control:} We propose Inference with Percentile Tuning, a mechanism that provides a monotonic conservativeness dial. This allows users to granularly manage the trade-off between overtriage and undertriage (the ROC operating point) without retraining.
  \item \textbf{Empirical Safety Preservation on Unambiguous Cases:} We demonstrate on two clinical benchmarks that STEER achieves broader behavioral coverage compared to temperature-based sampling and static persona ensembles. Compared to a representative post-training method, STEER maintains substantially higher accuracy on unambiguous urgent cases while providing comparable control over ambiguous decisions. 
\end{itemize}

\section{Related Works}
While the problem of mode collapse has spurred research across training, inference, and optimization, existing solutions largely treat diversity as a static objective rather than a dynamic, steerable operational parameter. We depart from these paradigms by reframing the goal: we do not solely seek to restore the distribution, but to instrument it for consistent human control.

\textbf{Training-Time Distributional Alignment.} Attempts to mitigate mode collapse most often intervene during post-training. Spectrum Tuning tunes base models on curated distributional tasks to better approximate valid output spaces \cite{Sorensen2025spectrum}. Similarly, methods like DivPO and DARLING explicitly modify the RLHF objective, integrating diversity rewards to penalize semantic redundancy \cite{Lanchantin2025-ij, Li2025-ft}. However, these solutions face  limitations. First, their requirement to modify model parameters limit them to open-weight models. This renders them inapplicable to the many powerful proprietary systems that are most frequently deployed in high-stake settings. Second, they produce static artifacts with rigid learned distributions. Adapting to a shifting risk profile (e.g., pandemic vs. normal operations) requires expensive retraining rather than simple inference-time reconfiguration. Third, global distributional updates can inadvertently degrade performance on unambiguous inputs, exhibiting a form of catastrophic forgetting where the model unlearns fundamental knowledge in its pursuit of diversity. Finally, training-time methods require the curation of massive multi-task datasets (e.g., the 40-source SPECTRUM SUITE) and significant compute for parameter updates. STEER circumvents these limitations entirely. As a model-agnostic framework, it applies to any SOTA model and offers superior flexibility, allowing users to define and swap objective functions without modifying model weights.

\textbf{Inference-Time Elicitation and Control.} To avoid the costs of retraining, recent work has explored prompting strategies to elicit latent diversity. Verbalized Sampling (VS) instructs models to articulate probabilities over a list of potential answers \cite{Zhang2025-pn}. However, VS maximizes diversity blindly, expanding the output space without guaranteeing that the generated options cover the specific risk gaps relevant to safety-critical decision-making. Crucially, it lacks a mechanism for actionable selection. It presents ranges of options but provides no policy to navigate them. STEER advances beyond this by coupling elicited diversity with Inference with Percentile Tuning, transforming the chaotic distribution into an ordered risk spectrum. STEER distinguishes from SteerConf, which employs semantic steering (e.g., "be very cautious") but for a fundamentally different objective of confidence calibration \cite{Zhou2025-rk}. SteerConf utilizes the ensemble to triangulate the single most reliable answer and assign it a calibrated probability score, effectively treating ensemble variance as uncertainty to be resolved. In contrast, STEER treats this variance as a valid policy landscape to be navigated. Rather than collapsing the ensemble to a single truth based on internal certainty, STEER maps the ensemble's outputs to ordinal levels of care. This allows the user to select a decision based on external risk tolerance via our conservativeness dial, a capability that current methods do not support.

\textbf{Evolutionary Optimization and Multi-Agent Systems.} The fundamental distinction between STEER and existing persona-based frameworks lies in the optimization objective. Multi-agent debate \cite{Du2023-db, Liang2023-rj, Suzgun2024-qb} and prompt evolution \cite{Agrawal2025-gj, Fernando2023-qw, Guo2023-mt, Yang2023-nr} methods incentivize consensus or optimize for a single best prompt or answer. This effectively fixes the model to a static operating point, failing to provide the continuous control required for dynamic environments. STEER reverses this paradigm by adopting a quality-diversity perspective (maximizing valid coverage) \cite{Qian2024-cn}. Rather than converging on one optimal agent, STEER utilizes operators to evolve a population that explicitly maximizes diversity. This ensures the ensemble spans the full ordinal risk spectrum, transforming the goal from maximizing accuracy to maximizing valid operational range.

\section{Problem Formulation}
\label{sec:problem_formulation}

While validated in clinical triage, our framework applies to any ordinal decision process (e.g., credit risk, legal evaluation). We formalize the challenge not as classification, but as a constrained search for behavioral diversity.

\subsection{Latent Context and Acceptable Disagreement}
Let $x$ denote an input case and $y \in \mathcal{Y}$ an ordinal decision (e.g., triage level 1--5 representing urgency level of care). For ambiguous inputs, there exists a latent distribution $P(y|x)$ over acceptable decisions induced by expert disagreement and contextual uncertainty. While this distribution is not directly observable, its existence motivates methods that preserve decision diversity rather than collapsing to a single mode. Standard alignment techniques optimize toward a singular reference, effectively treating this valid disagreement as noise. We define mode collapse as the loss of this latent variance, rendering models incapable of representing professional disagreement or adapting to shifting risk thresholds.

\subsection{Defining Behavioral Diversity}
\label{subsec:diversity_def}
We formally define the diversity we seek to recover. Let the persona pool be $\mathcal{P}$. Each rater persona $j \in \mathcal{P}$ produces an ordinal decision $RC(i,j) \in \{1,\dots,K\}$ for case $i$, where $K$ indexes the discrete ordinal decision categories. For each case $i$, we define the empirical distribution over ordinal decisions induced by the persona pool as:
\begin{equation}
p_i(k) = \frac{1}{|\mathcal{P}|} \sum_{j \in \mathcal{P}} \mathbb{I}[RC(i,j) = k], \quad k = 1,\dots,K
\end{equation}
\noindent where $\mathbb{I}$ is the indicator function. We quantify behavioral diversity using the normalized per-case decision entropy:
\begin{equation}
\tilde{H}(p_i) = -\frac{1}{\log K} \sum_{k=1}^K p_i(k)\log p_i(k)
\end{equation}
\noindent Let $\mathcal{A}$ denote the subset of ambiguous cases. We define the global diversity objective $\mathcal{D}(\mathcal{P})$ as the mean entropy across these ambiguous inputs:
\begin{equation}
\label{eq:mean_entropy}
\mathcal{D}(\mathcal{P}) = \frac{1}{|\mathcal{A}|} \sum_{i \in \mathcal{A}} \tilde{H}(p_i)
\end{equation}
$\mathcal{D}(\mathcal{P})$ is high when, on ambiguous cases, the personas distribute decisions across multiple valid ordinal levels, and low when the population collapses to a single response. We emphasize that $D(P)$ is defined purely in terms of observed model outputs and does not encode safety, validity, or correctness.

\subsection{Constrained Quality-Diversity Search}
\label{subsec:qd_objective}

We frame the problem as a quality-diversity search. We seek to find a population $\mathcal{P}$ that maximizes the diversity objective $\mathcal{D}(\mathcal{P})$ subject to strict feasibility constraints on every individual persona $j$. Let $S(j)$, $C(j)$, and $V(j)$ denote the safety, coherence, and variance scores for a persona (defined formally in Section \ref{subsec:constraints}). 

We define the set of feasible personas $\Phi$ as those meeting minimum thresholds $\tau$:
\begin{equation}
\label{eq:qd_objective}
\Phi = \{j \mid S(j) \ge \tau_{safe} \land C(j) \ge \tau_{coh} \land V(j) \le \tau_{var}\}
\end{equation}
Our objective is to discover a population $\mathcal{P} \subseteq \Phi$ that maximizes Eq. 3. This formulation ensures that we maximize behavioral coverage strictly within the bounds of safety, reasoning validity, and signal stability. Note that to maintain population health during evolution, these thresholds $\tau$ are implemented as adaptive percentiles rather than fixed absolute values (Appendix \ref{app:hyperparameters}).

\subsection{Persona Bias Model}
To systematically navigate the search space to maximize $\mathcal{D}(\mathcal{P})$, we require a low-dimensional embedding of each persona's behavior. We utilize a Maximum Likelihood Estimation (MLE) formulation, treating the discrete ordinal labels $y \in \{1, \dots, K\}$ as integer scores to enable continuous parameter estimation. We model the observed urgency $RC(i,j)$ for case $i$ by persona $j$ as:\begin{equation}
\label{eq:rc_model}
RC(i, j) = \Theta_i + u_j + \epsilon_{ij} 
\end{equation}
\noindent where $\Theta_i$ represents the latent case-specific difficulty and $u_j$ is the scalar bias parameter for the persona. The term $\epsilon_{ij}$ captures the residual variance ($s_j^2$), representing the degree to which a persona's behavior deviates from a consistent linear bias trend. Here, $u_j$ serves as the behavioral descriptor for persona $j$. While entropy (Eq. \ref{eq:mean_entropy}) is the global objective, $u_j$ is the coordinate used by the evolutionary algorithm to identify gaps and target specific regions of the risk spectrum. We acknowledge that modeling ordinal data as continuous variables is a deliberate simplification. However, our goal is not to fully model expert cognition or calibrate density estimation, but to construct an ordered ensemble that spans a monotonic axis of conservativeness. Treating labels as integers provides a robust proxy for this ranking purpose. To ensure model identifiability (resolving the shift-invariance between $\Theta$ and $u$), we enforce a centering constraint $\sum_{i} \Theta_i = 0$ during optimization.

\subsection{Regularization Constraints}
\label{subsec:constraints}
To prevent the system from achieving maximal entropy via random noise (hallucinations) or stochastic instability, we define the constraint functions from Eq. \ref{eq:qd_objective}. While these constraints can be customized to domain-specific needs (e.g., adding fairness or tone constraints), for our clinical application we define:
\begin{itemize}
    \item \textbf{Safety Constraint $S(j)$:} Ensures ground-truth alignment on unambiguous inputs. The ensemble must maintain high accuracy ($\ge \tau_{\text{safe}}$) on cases where consensus exists, preventing the model from losing fundamental domain knowledge in blind pursuit of diversity.
    \item \textbf{Coherence Constraint $C(j)$:} Ensures reasoning validity. Generated rationales must rely on present evidence rather than hallucinated antecedents, enforcing a minimum standard of logical soundness.
    \item \textbf{Variance Constraint $V(j)$:} Ensures signal stability. We minimize intra-persona inconsistency, estimated via the residual variance ($s_j^2$) of the persona's outputs relative to the linear bias model. This penalizes personas that behave erratically (noisy residuals) rather than following a consistent risk posture.
\end{itemize}

\subsection{Control Mechanism: The Ordinal Performance Curve}
Finally, we define the steerability of the system via the ordinal performance curve. This metric plots overtriage vs undertriage rates as the selection parameter $P$ sweeps from 0 to 100. While a static model represents a single fixed point on this 2D plane, our framework aims to generate the frontier, enabling operators to dynamically select an operating point that satisfies desired tolerances.

\section{Method: The STEER Framework}
\label{sec:method}

\begin{figure*}[ht] 
    \centering
    \includegraphics[width=0.9\textwidth]{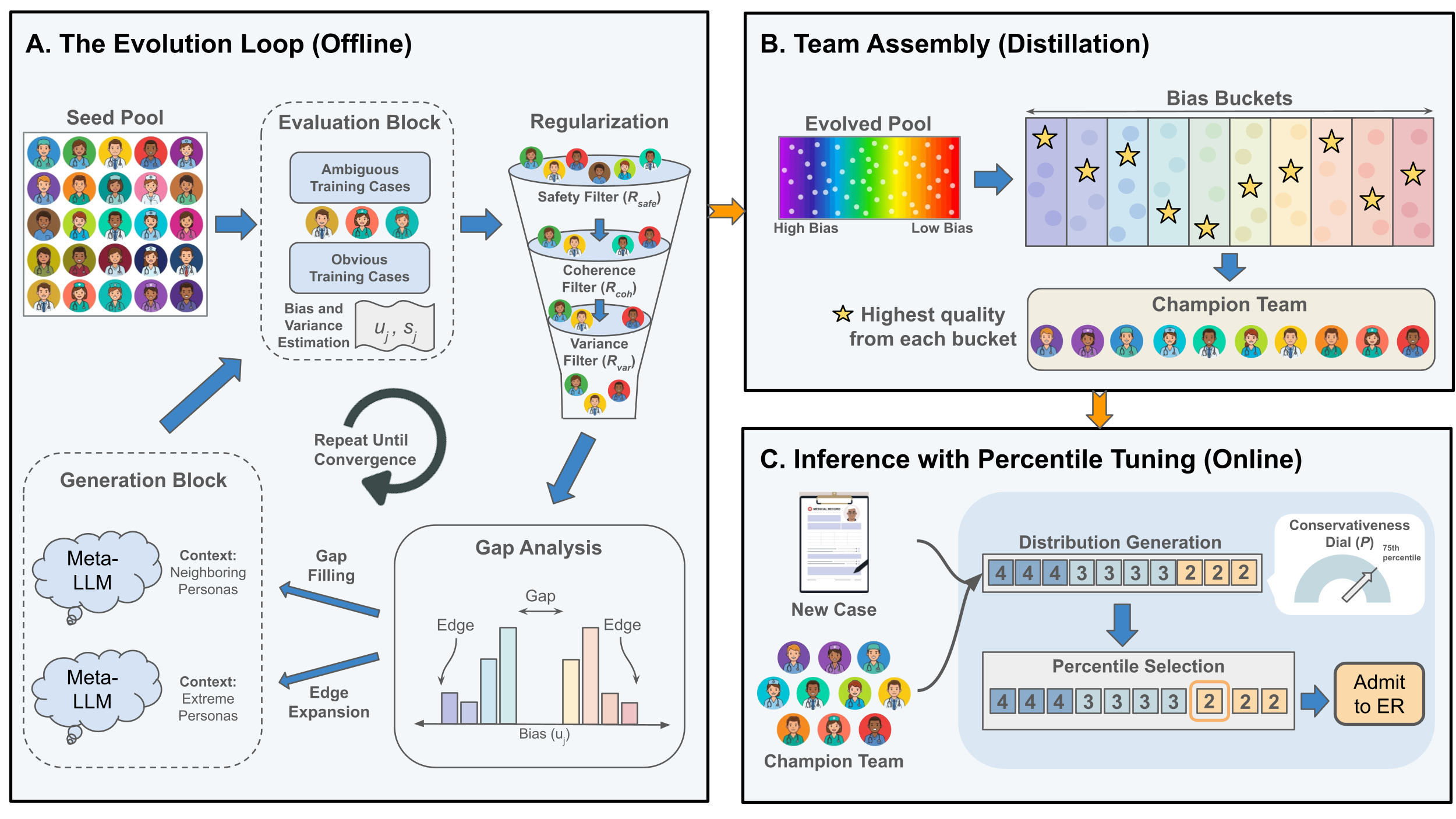}
    \caption{\textbf{The STEER Framework. (A) Evolution:} We iteratively optimize a persona population to maximize bias diversity subject to constraints. \textbf{(B) Assembly:} The evolved pool is distilled into a compact team to ensure uniform spectral coverage. \textbf{(C) Inference:} At deployment, the team acts as a distributional generator. The Inference with Percentile Tuning mechanism acts as a conservativeness dial.}
    \label{fig:steer_framework_overview}
\end{figure*}

STEER comprises an offline evolutionary search for a persona ensemble and an online inference mechanism (Figure \ref{fig:steer_framework_overview}). The offline evolution enables low-latency inference.

\subsection{Phase 1: Persona Evolution via Constrained Quality-Diversity Search}

We construct the persona ensemble using an iterative evolution procedure that progresses through evaluation, selection, and generation. This loop continues until a compute budget or a convergence metric (e.g. coverage plateau) is met. We explicitly frame this as a heuristic search designed to trade formal optimality guarantees for practical applicability to black-box models.

\subsubsection{Initialization}
The process begins with an initial seed persona pool. This step offers a key opportunity to inject domain knowledge, where users may seed the pool with known expert archetypes (e.g., ``risk-averse clinician``, ``efficiency-focused triage nurse``) or start with a smaller, generic set to rely more heavily on the algorithm's discovery capabilities. 

\subsubsection{Evaluation}
In each generation, every active persona is evaluated on ambiguous cases and unambiguous cases to compute three descriptors. First, we aggregate the ordinal outputs ($RC$) and apply the Maximum Likelihood Estimation (MLE) model (Eq. \ref{eq:rc_model}). We fit the parameters $\Theta$ and $u$ by minimizing the mean squared error loss $\mathcal{L} = \sum_{i,j} (RC_{obs}^{(i,j)} - (\Theta_i + u_j))^2$ via PyTorch and the Adam optimizer \cite{Kingma2014-uy}, subject to the identifiability constraint $\sum \Theta_i = 0$. The variance parameter $s_j^2$ is subsequently calculated as the mean squared residual of persona $j$'s predictions, serving as a proxy for behavioral inconsistency. Second, we compute a reasoning coherence score to penalize hallucinations or logical disconnects. While we implement this via an LLM-as-judge for scalability in our experiments, we emphasize that this is a modular slot for any domain-appropriate reward signal (e.g., rule-based logic checks or human feedback) and is not intrinsic to the optimization algorithm itself. Third, we calculate a safety score by measuring strict accuracy specifically on the subset of unambiguous cases.

\subsubsection{Selection}
\label{subsec:selection}
We apply feasibility constraints (Sec \ref{subsec:qd_objective}) sequentially. The safety constraint ($S(j) \ge \tau_{safe}$) discards personas failing a strict accuracy threshold on unambiguous cases, ensuring the ensemble maintains domain competence. The coherence constraint ($C(j) \ge \tau_{coh}$) prunes the bottom tier of the population based on LLM-judge reasoning scores. Finally, the variance constraint ($V(j) \le \tau_{var}$) removes high-variance personas within bias-clustered groups (see Appendix B.5) to distinguish systematic bias from random noise.

\subsubsection{Generation}
To discover new behaviors, we employ two operators: (1) Gap Filling, where the meta-LLM generates a persona between two neighbors to bridge discontinuities in the sorted bias list; and (2) Edge Expansion, where it generates valid personas more conservative or aggressive than the current population extremes to extend the dynamic range.

\subsection{Phase 2: Final Team Assembly}
Once the evolution loop terminates, we distill the large optimized pool into a compact champion team for deployment. This step is motivated by practical efficiency. While a massive ensemble maximizes theoretical coverage, the marginal gain in diversity diminishes rapidly with team size, whereas the computational cost of inference scales linearly. To approximate the full spectrum's coverage using $N$ personas (corresponding to a budget of $N$ inferences per case), we employ a coverage-based selection algorithm. The assembly  begins by anchoring the team with the most conservative and aggressive personas from the pool to preserve the full operational dynamic range. We then partition the remaining bias spectrum between these extremes into equal-width intervals corresponding to the remaining team slots. Within each bucket, we select a single representative persona by prioritizing candidates first by their safety and coherence scores, and then by minimizing variance. In the event of a sparse region where a bucket is empty, the algorithm selects the highest-quality available persona nearest to that bucket’s center. Alternatively, one can use specialized set cover algorithms, which we leave as future work.

\subsection{Inference with Percentile Tuning}
During deployment, the final team functions as a distributional generator, where $N$ personas produce distinct predictions for a given case. Because the evolutionary search is offline and these inference queries are fully parallelizable, STEER maintains latency comparable to a single model call. To translate this distribution into a single actionable decision, we apply Inference with Percentile Tuning. This deterministic mechanism first sorts the $N$ outputs by urgency magnitude (e.g., from "Self-care" to "Emergency"). A user-specified parameter $P \in [0, 100]$ (the conservativeness dial) is then mapped to a rank index $k = \lfloor \frac{P}{100} \times (N-1) \rfloor$, and the system outputs the decision at rank $k$. Because the outputs are sorted, this function guarantees monotonicity, where increasing $P$ strictly results in an output that is either equal to or more conservative than the previous state.

\section{Experimental Setup}

We evaluate STEER on two real-world clinical decision-making tasks: a public benchmark for reproducibility and a private EHR dataset to test generalizability. 

\subsection{Datasets}
\textbf{MIETIC (MIMIC-IV-Ext Triage Instruction Corpus).} \cite{Shen2025-xc} To evaluate performance in emergency department triage, we utilize MIETIC, a structured dataset of 9,629 expert-validated triage cases derived from MIMIC-IV \cite{Johnson2024-us}. Each case includes chief complaints, vital signs, demographics, and medical history, labeled with the Emergency Severity Index (ESI) from 1 (most urgent) to 5 (least urgent). We use ESI-3 (Moderate Risk) as the ambiguous set for evolution ($N=1028$) and ESI-1/5 as the unambiguous safety set ($N=2218$) to enforce constraints.

\textbf{Real-World EHR Symptom Triage.} To assess generalization, we utilize a de-identified, proprietary dataset curated from real-world Electronic Health Records (EHR) containing 560 patient cases. This dataset focuses on symptom urgency assessment in a telehealth context. Each case aggregates a comprehensive patient context, consisting of: (1) a patient summary and last encounter notes; (2) retrieved EHR structured data (medications, recent labs, conditions); and (3) a multi-turn conversation between a patient simulator and an AI physician agent established in prior work. The output space consists of a 5-point ordinal urgency scale ranging from emergency/urgent care (Level 1) to self-care (Level 5). Ground truth was established by three senior human physicians. Cases with unanimous consensus on the extremes (Level 1 or 5) constitute the unambiguous split (N=109). All other cases, characterized by inter-annotator disagreement or intermediate urgency, constitute the ambiguous split (451). We partition each dataset into 70/15/15 training, validation, and testing splits (see Appendix \ref{app:dataset_details}).

\subsection{Models and Baselines}

We conduct two sets of experiments to isolate the impact of inference-time steering versus post-training alignment. 

\textbf{Experiment I: Evolutionary Discovery with Closed-Source Models.}
The evaluation demonstrates how STEER unlocks distributional capabilities in SOTA proprietary models where weight access is impossible. We use GPT-5-mini as the base model \cite{openai2025gpt5}. To ensure a fair comparison, we hold the compute budget constant and use identical system prompts across all methods, varying only the injected persona instructions  (Appendix \ref{app:algorithm_details}).

First, we evaluate a high-temperature sampling baseline, where the base model is queried 10 times per case with temperature $T=1.0$ to elicit stochastic diversity. Second, we evaluate a static persona baseline, consisting of 10 distinct personas randomly sampled from the same expert seed pool used to initialize evolution (Appendix \ref{app:seed_personas}). This comparison isolates the specific value of evolutionary optimization over human-curated heuristics. For our proposed method, we first evolve a large pool of 75 personas using the STEER algorithm. To analyze the efficiency-diversity trade-off, we then distill this pool into champion teams of varying sizes ($N \in \{5, 10, 20, 30\}$). Finally, we apply the Inference with Percentile Tuning mechanism across the outputs of all three methods (high-temp, static, and STEER) to generate and compare their respective performances.

\textbf{Experiment II: Inference vs. Post-Training.}
To compare our inference-time framework against a representative post-training intervention, we utilize the open-source Gemma family of models \cite{team2024gemma}. In this experiment, we apply STEER (20 evolved personas + Percentile Tuning) to the standard instruction-tuned Gemma-3-12B-IT model. We benchmark this against the pre-trained Spectrum-Tuned version of the same base model \cite{Sorensen2025spectrum}. Both models utilize the same base prompt structure (Appendix \ref{app:algorithm_details}). We utilize the general-purpose checkpoint fine-tuned on distributional tasks to assess off-the-shelf steerability. We generate 20 outputs per case using standard temperature sampling ($T=1.0$) to elicit the model's learned diversity. To isolate the impact of the distributional source (inference-time personas vs. post-training weights), we apply Percentile Tuning to the outputs of both models. This ensures strictly comparable steerability mechanisms.

\subsection{Evaluation Metrics}
Because ambiguous cases (Level 3) lack a singular ground truth, we define performance relative to the  midpoint. Overtriage is defined as estimating higher urgency ($\hat{y} < 3$), while undertriage is estimating lower urgency ($\hat{y} > 3$). The ordinal performance curve plots the trade-off between the overtriage rate and the safe rate ($1 - \text{undertriage rate}$), generated by sweeping the control parameter $P \in [0, 100]$. We quantify the valid operating region via the ordinal AUC, calculated using right-step integration $\sum (x_{i+1} - x_i) y_i$.

\section{Results}
We demonstrate that STEER: (1) significantly enhances operational range vs. baselines; (2) provides deterministic control over the ROC point; and (3) prevents the catastrophic forgetting seen in post-training. Error bars in all figures represent 95\% bootstrap confidence intervals calculated via resampling test cases ($n=2000$ iterations).

\subsection{Steerability in Closed-Source Models}
\label{subsec:results_closed_source}
\begin{figure}[ht] 
    \centering
    \includegraphics[width=0.45\textwidth]{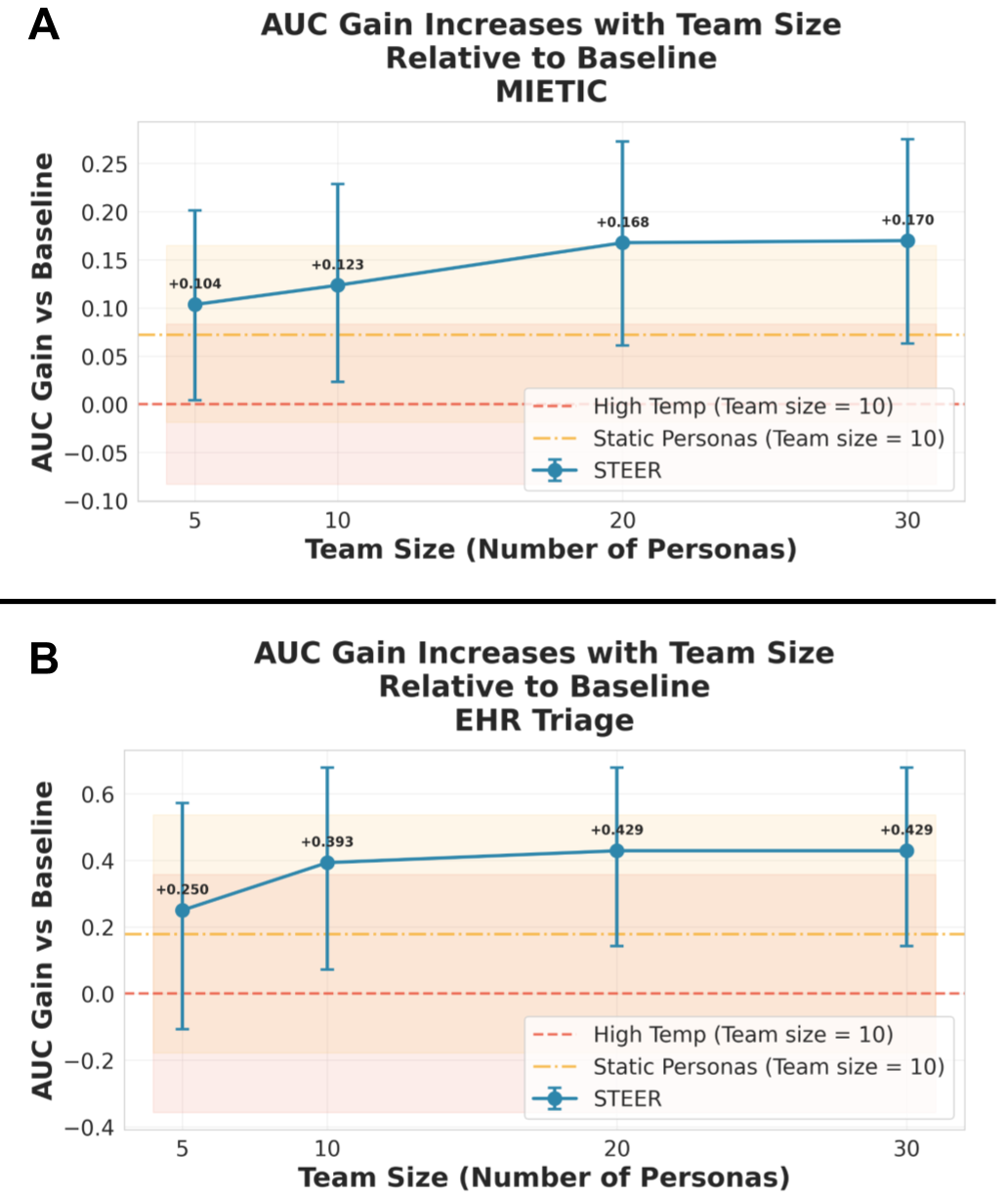}
    \caption{\textbf{Evolutionary Steerability vs. Baselines.} Ordinal AUC gain of STEER ensembles (varying size $N \in \{5, 10, 20, 30\}$) relative to high-temperature sampling ($N=10$) and static persona baselines ($N=10$) for \textbf{(A) MIETIC Triage} and \textbf{(B) EHR Triage.}}
    \label{fig:auc_gains}
\end{figure}
Figure \ref{fig:auc_gains} illustrates that STEER consistently outperforms both baselines in Ordinal AUC, enhancing output space coverage. On the EHR Triage task, STEER achieves an AUC gain of $0.429$ over the baseline, nearly double the static team's performance. The evolutionary process successfully discovers non-intuitive bias gaps that manual curation misses. Additionally, we observe diminishing returns with respect to team size. A team size of $N=10$ captures the majority of the diversity gain (e.g., $+0.393$ on EHR Triage) while incurring only 33\% of the inference cost of the $N=30$ team, confirming the efficiency of our coverage-based assembly selection. This allows practitioners to select the optimal deployment configuration for resource-constrained environments.

\subsection{Deterministic Control via Percentile Tuning}
\label{subsec:results_percentile}

\begin{figure}[ht] 
    \centering
    \includegraphics[width=0.40\textwidth]{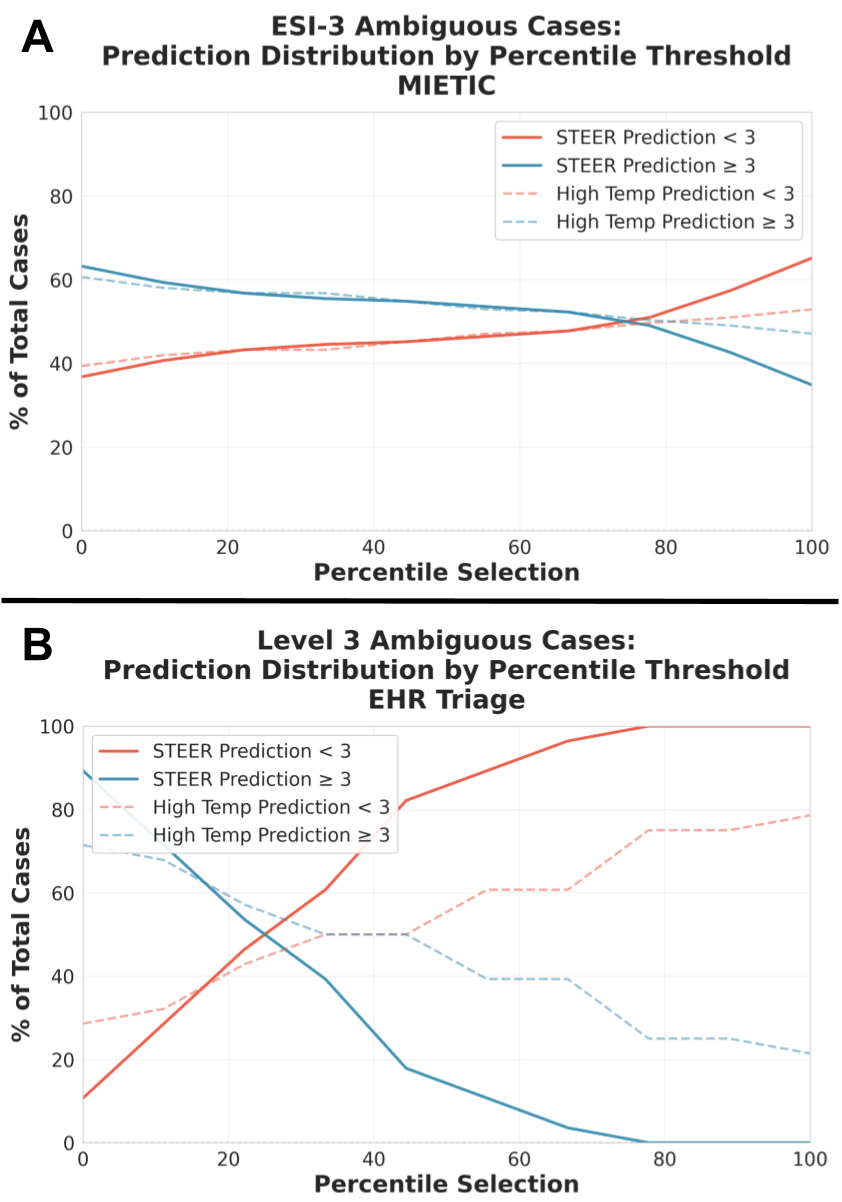}
    \caption{\textbf{Deterministic Risk Control via Percentile Tuning.} The effect of the Percentile Selection dial on the predicted urgency distribution for STEER ($N=10$) vs. the high-temperature baseline ($N=10$) on \textbf{(A) MIETIC} and \textbf{(B) EHR Triage}. Higher percentiles ($P \to 100$) correspond to strictly more conservative (urgent) predictions.}
    \label{fig:percentile_tuning}
\end{figure}
We next evaluate the controllability of the expanded range. Figure \ref{fig:percentile_tuning} visualizes the impact of the conservativeness dial on the predicted case distribution. An operator increasing the conservativeness must be guaranteed that the model behavior will shift in one direction. STEER's Percentile Tuning mechanism enforces this by sorting the ensemble's outputs. Increasing the percentile $P$ results in a strictly non-decreasing shift in urgency. The proportion of cases predicted as high-acuity rises while low-acuity predictions fall, eliminating erratic behavior associated with standard prompt engineering. Notably, the high-temperature sampling baseline and STEER differ significantly in sensitivity—the magnitude of change elicited by the dial. The baseline exhibits a constrained dynamic range, where the error rates remain relatively static regardless of the percentile selected. This highlights mode collapse, where the model's internal variance is too narrow to be effectively steered. In contrast, STEER unlocks a wide operating window. For EHR triage, by shifting the dial from $P=20$ to $P=80$, the system reduces the risk of undertriage from $\approx 60\%$ to $<5\%$, granting the operator meaningful control over the safety profile. For MIETIC, STEER marginally widens the decision boundaries compared to the  baseline. The STEER curves achieve a lower undertriage rate at high percentiles, proving more responsive for tuning the system to specific conservativeness preferences.

\subsection{Inference-Time vs. Post-Training 
Alignment}
\label{subsec:results_training_comparison}
\begin{figure}[ht] 
    \centering
    \includegraphics[width=0.48\textwidth]{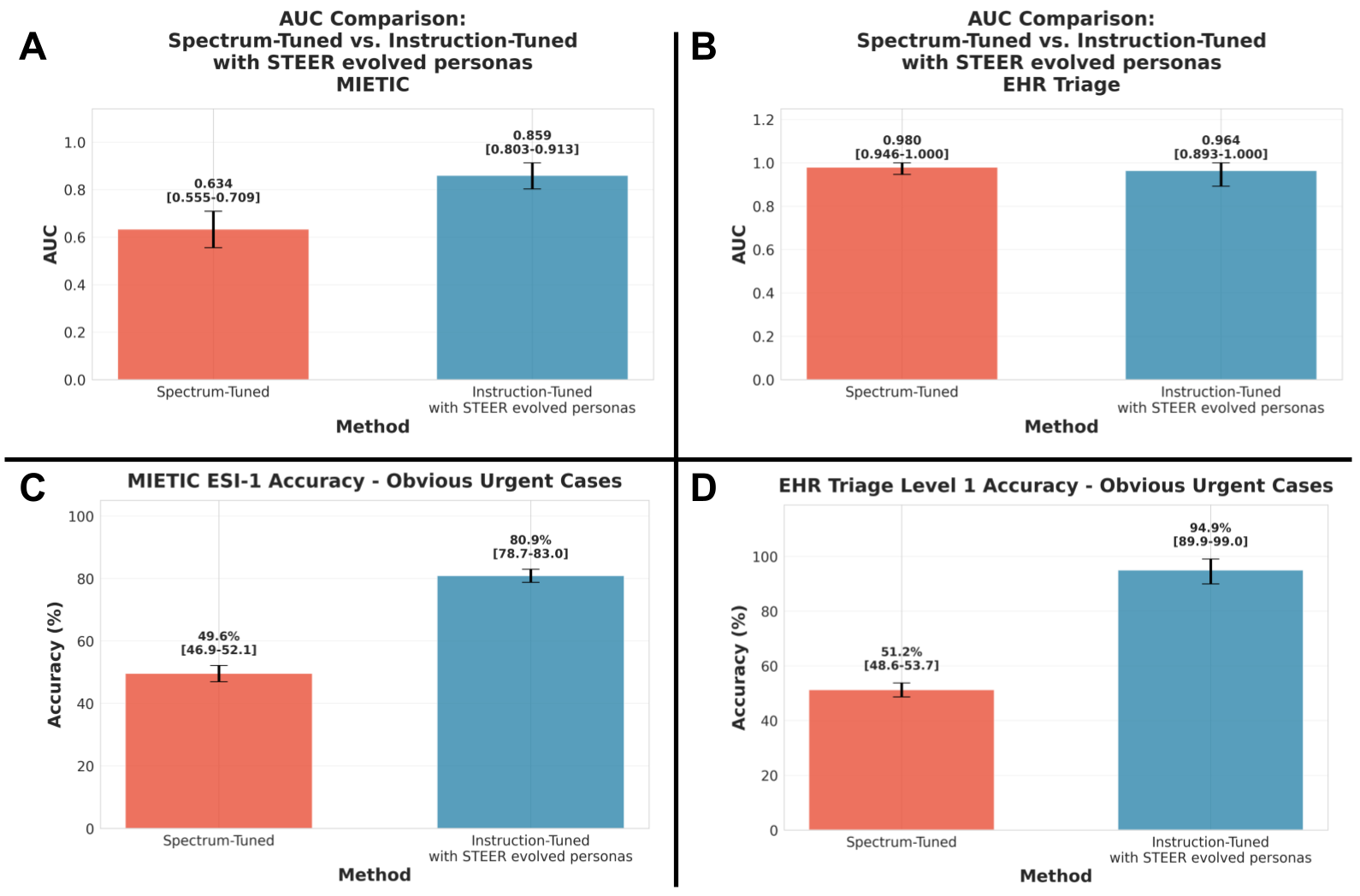}
    \caption{\textbf{Inference-Time Evolution vs. Post-Training Alignment.} Comparison between Spectrum Tuning (red) and STEER (blue) on Gemma-3-12B-IT. \textbf{(A, B) Steerability:} Ordinal AUC scores on the MIETIC and EHR Triage datasets. \textbf{(C, D) Safety:} Accuracy on the unambiguous urgent safety sets.}
    \label{fig:spectrum_comparison}
\end{figure}
Finally, we compare STEER against Spectrum Tuning, a representative post-training method. We evaluate whether enhancing the output space coverage via post-training is superior to improving it via STEER. As shown in Figure \ref{fig:spectrum_comparison} (Panels A \& B), STEER applied to a standard instruction-tuned model achieves Ordinal AUC scores highly competitive with the Spectrum-Tuned model (0.859 vs. 0.634 on MIETIC; 0.964 vs. 0.980 on EHR). This result shows that inference-time evolution can sufficiently enhance valid output space. Crucially, Panels C and D reveal a failure mode in the post-training approach. The Spectrum-Tuned model suffers significant degradation on unambiguous urgent cases assessed for safety, dropping to $49.6\%$ accuracy on MIETIC and $51.2\%$ on EHR. This indicates that Spectrum Tuning reduces the model's domain competence to increase diversity. In contrast, STEER maintains high safety performance ($80.9\%$ and $94.9\%$ respectively). This robustness is a consequence of our constrained optimization objective ($R_{safe}$), which allows STEER to pursue diversity without sacrificing reliability on critical, unambiguous cases.

\section{Conclusion}

Current alignment paradigms optimize for average correctness, inadvertently collapsing valid professional disagreement into a single point estimate. This renders models rigid, stripping them of the ability to trade off sensitivity and specificity (the ROC operating point) based on changing environmental constraints.

In this work, we introduced STEER, a framework that restores this control by transforming frozen, black-box models into tunable instruments. By reformulating steerability as a constrained Quality-Diversity search, we demonstrated that it is possible to expand the latent spectrum of valid risk perspectives without expensive retraining. Our evolutionary approach successfully discovers bias gaps, while our Percentile Tuning mechanism provides the deterministic monotonicity required for safety-critical deployment. Crucially, our results show that this inference-time intervention matches the steerability of state-of-the-art post-training methods while preventing the catastrophic forgetting of fundamental safety knowledge on unambiguous cases. This suggests a new paradigm for reliable AI, where reliability is defined not by a single static output, but by the system's ability to be calibrated to the user's operational tolerance.

Our work opens clear avenues for future investigation. First, while STEER expands the valid spectrum, the computational cost of ensemble inference scales linearly with team size. Future work could explore distillation techniques to compress the diverse team into a single, conditionally steerable adapter. Second, our current evaluation focused on ordinal decision tasks. Expanding the framework's constraints and tuning mechanisms to support open-ended generation remains an active challenge. Finally, while we demonstrate utility in clinical triage, the framework's core principle of managing risk via calibrated diversity is broadly applicable to legal, financial, and ethical reasoning, warranting further validation in these domains.

Ultimately, STEER represents a step toward pluralistic AI systems that do not merely dictate a single "correct" answer, but empower human experts with a transparent, controllable range of valid options.

\section*{Impact Statement}

\textbf{Promoting Pluralism and Context-Aware Safety.} Standard alignment techniques optimize for average correctness, artificially collapsing valid professional disagreement into a single "safe" mode. By employing quality-diversity search to recover the latent spectrum of expert opinion, our framework enables context-aware safety. This allows AI systems to adapt their sensitivity/specificity trade-off (the ROC operating point) to dynamic environmental constraints rather than imposing a rigid risk posture that may be locally disastrous. This is particularly impactful for democratizing high-quality decision support in under-resourced environments where retraining models is infeasible.

\textbf{Ethical Risks of Tunable Efficiency.} The introduction of a deterministic risk dial introduces friction between clinical safety and economic efficiency. There is a non-trivial risk that operators could utilize the dial to minimize overtriage strictly for cost-saving purposes, effectively automating undertriage to an ethically unacceptable level. However, we argue that current black-box models effectively make this decision implicitly (and rigidly). STEER makes this trade-off transparent. Nonetheless, deployment must be accompanied by strict governance protocols where the operating point ($P$) is treated as a clinical calibration parameter governed by medical directors, rather than a user preference.

\textbf{Stability of Inference-Time Governance.} Because STEER functions as an external wrapper around a frozen base model, safety becomes a runtime dependency rather than a static weight property. This creates a risk of alignment drift. If the underlying proprietary model is updated, quantized, or (in the case of APIs) silently patched, the ensemble’s evolved calibration may degrade. Our inference-time approach requires continuous monitoring and potential re-evolution to maintain its safety guarantees against API shifts.


\bibliography{example_paper}
\bibliographystyle{icml2026}

\newpage
\appendix
\onecolumn
\section{Dataset Details \& Statistics}
\label{app:dataset_details}

\subsection{MIETIC (Public Benchmark)}
\label{app:mietic_details}

\textbf{Preprocessing and Input Formatting.} The MIETIC dataset is derived from MIMIC-IV emergency department notes \cite{Shen2025-xc, Johnson2024-us}. Raw clinical notes sometimes contain information that would only be known \textit{after} triage (data leakage). To address this, we preprocessed the notes into a single narrative string containing only the ``Chief Complaint,'' ``Vitals,'' and ``Medical History.'' We employed a separate LLM ((\textbf{GPT-5-mini}) to perform information extraction, strictly retaining information that the patient could reasonably have known or presented with \textit{before} being seen by a physician.

\textbf{Synthetic Input Example.} Below is an illustrative example of the preprocessed input format. \textit{Note: This is a synthetic case generated to mirror the style and structure of the dataset while protecting patient privacy. It does not represent a real individual.}

\begin{quote}
\small
\textbf{Input String:} ``A 72-year-old male with a history of COPD and chronic heart failure presents to the ED reporting worsening shortness of breath and fluid retention over the past 3 days. He notes increased swelling in lower extremities and difficulty sleeping flat. On arrival, vitals were BP 135/85, HR 110, RR 28, and SpO2 91\% on room air. PMH includes hypertension, hyperlipidemia, and type 2 diabetes. Patient arrived via private vehicle.''
\end{quote}

\textbf{Class Distributions.} We utilized specific Emergency Severity Index (ESI) levels to define our training and safety sets. ESI-3 serves as the source of ambiguous cases for evolution, while ESI-1 (Most Urgent) and ESI-5 (Least Urgent) serve as safety constraints. The demographic and clinical characteristics of these subsets are detailed below.

\begin{table}[H]
\centering
\caption{Demographic Statistics (MIETIC)}
\label{tab:mietic_demographics}
\begin{small}
\begin{tabular}{lrrr}
\toprule
\textbf{Statistic} & \textbf{ESI-1 (Critical)} & \textbf{ESI-3 (Ambiguous)} & \textbf{ESI-5 (Non-Urgent)} \\
\midrule
\textit{Count ($N$)} & 955 & 1028 & 1263 \\
\midrule
\textbf{Gender} & & & \\
\hspace{3mm} Male & 530 (55.5\%) & 428 (41.6\%) & 748 (59.2\%) \\
\hspace{3mm} Female & 425 (44.5\%) & 598 (58.2\%) & 515 (40.8\%) \\
\hspace{3mm} Unknown & 0 (0\%) & 2 (0.2\%) & 0 (0\%) \\
\midrule
\textbf{Age (Years)} & & & \\
\hspace{3mm} Mean (SD) & 63.9 (17.8) & 45.6 (18.4) & 35.4 (14.5) \\
\hspace{3mm} Median & 66.0 & 44.0 & 29.0 \\
\hspace{3mm} Range & 18 -- 91 & 0 -- 91 & 18 -- 82 \\
\bottomrule
\end{tabular}
\end{small}
\end{table}

\begin{table}[H]
\centering
\caption{Distribution of Chief Complaints by ESI Level (MIETIC)}
\label{tab:mietic_complaints}
\begin{small}
\begin{tabular}{lrrr}
\toprule
\textbf{Category} & \textbf{ESI-1} & \textbf{ESI-3} & \textbf{ESI-5} \\
\midrule
Neurological & 263 & 73 & 2 \\
Respiratory & 218 & 82 & 19 \\
Gastrointestinal & 113 & 299 & 1 \\
Injuries & 110 & 131 & 315 \\
Cardiovascular & 89 & 45 & 26 \\
Constitutional & 67 & 44 & 13 \\
Musculoskeletal & 24 & 201 & 33 \\
Genitourinary & 10 & 75 & 0 \\
Dermatological & 18 & 29 & 239 \\
Poisoning & 15 & 4 & 4 \\
Sensory Organs & 3 & 5 & 93 \\
Psychological & 8 & 10 & 24 \\
Endocrine/Metabolic & 6 & 4 & 0 \\
Reproductive & 4 & 11 & 0 \\
Other / Unclassified & 7 & 15 & 494 \\
\bottomrule
\end{tabular}
\end{small}
\end{table}

\subsection{Real-World EHR Triage (Proprietary)}
\label{app:ehr_details}

\textbf{Dataset Construction.} This dataset consists of de-identified telehealth encounters. The input for each case is constructed by concatenating three distinct data sources into a single context string: (1) A generated patient summary and notes from the last encounter; (2) Structured EHR data including active medications, recent lab results, and problem lists; and (3) A multi-turn dialogue transcript between a patient simulator and an AI physician agent.

\textbf{Safety vs. Ambiguity Criteria.} Ground truth was established by three senior human physicians. We defined the subsets based on inter-annotator consensus:
\begin{itemize}
    \item \textbf{Safety Sets (Level 1 \& 5):} Cases where \textit{all three} annotators unanimously agreed on the extreme labels: ``Urgent care/Emergency'' (Level 1) or ``Self-care'' (Level 5).
    \item \textbf{Ambiguous Set (Level 3):} All remaining cases, which largely consist of disagreements (e.g., split decisions between PCP and ER), intra-rater ambiguity (e.g. multiple acceptable decisions within 1 annotator), or unanimous consensus on the intermediate ``Follow up with PCP'' label.
\end{itemize}

\textbf{Inter-Annotator Agreement.} The Fleiss' Kappa scores (Table \ref{tab:agreement}) highlight the inherent subjectivity of the task. While agreement is relatively higher for critical emergencies ($\kappa \approx 0.57$), it drops significantly for intermediate dispositions, confirming that ambiguity is a feature of the domain rather than label noise. 

\textbf{Agreement Metric Definitions.} To interpret the inter-annotator reliability reported in Table \ref{tab:agreement}, we define the calculated metrics as follows:

\begin{itemize}
    \item \textbf{Fleiss' Kappa:} A statistical measure for assessing the reliability of agreement between a fixed number of raters ($n=3$) when assigning categorical ratings. It corrects for agreement occurring by chance.
    \item \textbf{Avg Pairwise Kappa:} The arithmetic mean of Cohen's Kappa scores calculated for every unique pair of raters (Rater A vs. B, A vs. C, B vs. C). This provides a more granular view of pairwise consistency compared to the aggregate Fleiss' metric.
    \item \textbf{\% Unanimous Agreement:} The percentage of cases where all three annotators were in exact agreement (either all assigning the label or all rejecting it). This represents the rate of unambiguous consensus.
    \item \textbf{Assignment Rate:} The prevalence of the label within the dataset, calculated as the mean percentage of cases assigned that specific label across all annotators. This contextualizes the Kappa scores, as agreement metrics can be sensitive to highly imbalanced class distributions.
\end{itemize}

\begin{table}[H]
\centering
\caption{Human Annotator Agreement Statistics (EHR Triage)}
\label{tab:agreement}
\begin{small}
\begin{tabular}{lcccc}
\toprule
\textbf{Label Category} & \textbf{Fleiss' Kappa} & \textbf{Avg Pairwise $\kappa$} & \textbf{\% Unanimous} & \textbf{Assignment Rate} \\
\midrule
Self-care & 0.448 & 0.452 & 64.1\% & 31.7\% \\
Follow up with PCP & 0.310 & 0.340 & 49.3\% & 57.1\% \\
Urgent / Emergency & 0.569 & 0.570 & 86.3\% & 12.1\% \\
\bottomrule
\end{tabular}
\end{small}
\end{table}

\begin{table}[H]
\centering
\caption{Demographic Statistics (EHR Triage)}
\label{tab:ehr_demographics}
\begin{small}
\begin{tabular}{lrrr}
\toprule
\textbf{Statistic} & \textbf{Level 1 (Critical)} & \textbf{Level 3 (Ambiguous)} & \textbf{Level 5 (Self-Care)} \\
\midrule
\textit{Count ($N$)} & 76 & 451 & 33 \\
\midrule
\textbf{Gender} & & & \\
\hspace{3mm} Male & 19 (25.0\%) & 83 (18.4\%) & 7 (21.2\%) \\
\hspace{3mm} Female & 57 (75.0\%) & 368 (81.6\%) & 26 (78.8\%) \\
\midrule
\textbf{Age (Years)} & & & \\
\hspace{3mm} Mean (SD) & 50.5 (18.1) & 55.2 (18.6) & 58.4 (17.6) \\
\hspace{3mm} Median & 52.5 & 56.0 & 61.0 \\
\hspace{3mm} Range & 17 -- 95 & 17 -- 96 & 25 -- 96 \\
\bottomrule
\end{tabular}
\end{small}
\end{table}

\begin{table}[H]
\centering
\caption{Distribution of Chief Complaints by Care Level (EHR Triage)}
\label{tab:ehr_complaints}
\begin{small}
\begin{tabular}{lrrr}
\toprule
\textbf{Category} & \textbf{Level 1} & \textbf{Level 3} & \textbf{Level 5} \\
\midrule
Respiratory & 56 & 144 & 16 \\
Musculoskeletal & 7 & 107 & 1 \\
Genitourinary & 2 & 81 & 2 \\
Gastrointestinal & 3 & 36 & 3 \\
Sensory Organs & 2 & 22 & 0 \\
Injuries & 0 & 13 & 0 \\
Reproductive & 0 & 10 & 0 \\
Dermatological & 1 & 10 & 0 \\
Neurological & 1 & 9 & 2 \\
Cardiovascular & 0 & 4 & 9 \\
Constitutional & 3 & 7 & 0 \\
Other & 1 & 8 & 0 \\
\bottomrule
\end{tabular}
\end{small}
\end{table}

\section{Algorithm Implementation Details}
\label{app:algorithm_details}

\subsection{Inference Prompts}
\label{app:inference_prompts}

The final system prompts used during deployment differ slightly by dataset to align with the specific ordinal scale (ESI vs. Care Level). In both cases, the \texttt{\{persona\_block\}} is injected dynamically based on the specific agent persona being queried.

\subsubsection{System Prompt (MIETIC)}
\begin{quote}
\small
\texttt{- \{persona\_block\}}

\vspace{0.5em}
Your task is to assign an Emergency Severity Index (ESI) level (1–5) to the patient described below, consistent with both the case data and your overall approach.

First, carefully review the case details.
Next, use your clinical reasoning and reference the ESI definitions provided.
Let your perspective as described above guide how you interpret the information and weigh risks or resources.

\textbf{Patient Case:}
\texttt{\{patient\_case\}}

[... Standard ESI 1-5 Definitions Omitted for Brevity ...]

\textbf{Guidelines}
\begin{itemize}
    \item Count the type, not the number of individual tests (e.g., CBC + Lytes = 1 Lab Resource).
    \item Only include those resources most needed.
\end{itemize}

Think through your decision step by step, referencing both the case and your persona’s approach.
Respond only in the following JSON format:
\texttt{\{"reasoning": "...", "esi\_level": <int>\}}
\end{quote}

\subsubsection{System Prompt (HV Triage)}
\begin{quote}
\small
\texttt{- \{persona\_block\}}

\vspace{0.5em}
Your task is to assign a care level (1–5) to the patient described below, consistent with both the case data and your overall approach.

First, carefully review the case details.
Next, use your clinical reasoning and reference the care level definitions provided.
Let your perspective as described above guide how you interpret the information and weigh risks or urgency.

\textbf{Patient Case:}
\texttt{\{patient\_case\}}

[... Care Level 1-5 Definitions Omitted for Brevity ...]

Think through your decision step by step, referencing both the case and your persona's approach.
Respond only in the following JSON format:
\texttt{\{"reasoning": "...", "care\_level": <int>\}}
\end{quote}

\subsection{Initialization (Seed Personas)}
\label{app:seed_personas}
The evolutionary process was initialized with a small set of "seed" personas representing distinct, manually curated triage philosophies. These served as the genetic ancestors for the initial population. Below are a few illustrative examples:

\begin{itemize}
    \item \textbf{The Vital Signs Anchor:} You are a methodical triage nurse who begins every case by carefully reviewing the patient’s vital signs before considering history or symptoms. You believe vital stability provides the clearest initial snapshot of risk, and you use this as your anchor for further triage decision-making. Only after confirming stability do you weigh other presenting details for escalation.
    \item \textbf{The Narrative Interpreter:} You are a nurse who prefers a narrative approach, starting with the patient's story and description of symptoms. You find that understanding the timeline, progression, and patient concerns helps you contextualize clinical findings. You then map these qualitative details to the formal ESI criteria to arrive at your triage decision.
    \item \textbf{The Guideline Adherent:} You are an evidence-driven triage nurse who always consults hospital protocols and the latest ESI criteria before making a decision. Your workflow involves referencing each criterion in order, checking off whether each is met based strictly on documented data, and minimizing subjective interpretation. You value reproducibility and adherence to established guidelines above all.
\end{itemize}
For the Static Persona baseline used in Experiment I, we randomly sample $N$ personas from this same seed pool. This ensures that any performance gain observed in STEER is attributable to the evolutionary optimization of the population, rather than the quality of the initial prompt engineering.

\subsection{Evolutionary Prompts}
\label{app:evolutionary_prompts}

\subsubsection{Mutation Prompt (Gap Filling)}
To generate personas that interpolate between existing risk postures, the Meta-LLM (GPT-5-mini) is provided with "Reference Personas" (neighbors in the bias sort) and the following instructions:

\begin{quote}
\small
\textbf{Objective:} Create a realistic [Role] persona that would demonstrate triage behavior in a [Setting] with a bias estimate of approximately [Target Bias].

\textbf{Understanding the Bias Framework:}
\begin{itemize}
    \item \textbf{Negative bias:} Perceives cases as more urgent/serious than typical (Conservative).
    \item \textbf{Positive bias:} Perceives cases as less urgent than typical (Lenient).
\end{itemize}

\textbf{Target:} Create a persona whose triage philosophy and decision-making patterns would naturally produce bias $\approx$ [Target Bias].
\textbf{Context:} Filling gap between existing personas at [Start Bias] and [End Bias].

[...Reference Personas Inserted Here...]

\textbf{Generation Guidance:} Consider weaving together elements like:
\begin{itemize}
    \item \textbf{Formative Events:} What experiences shaped their approach to uncertainty?
    \item \textbf{Clinical Philosophy:} How do they balance caution vs. efficiency?
\end{itemize}
\textbf{Requirements:} Format as second person ('You are...'), 3-5 sentences.
\end{quote}

\subsubsection{Edge Expansion Prompt}
To push the boundaries of the ensemble's operational range, the Meta-LLM (GPT-5-mini) is instructed to extrapolate beyond the current extremes:

\begin{quote}
\small
\textbf{Objective:} Create a [Role] with extreme but realistic triage patterns in a [Setting].
\textbf{Target bias:} [Target Bias] (More [Conservative/Lenient] than current extreme).

\textbf{Context:} This persona represents an outlier at the edge of the bias distribution—significantly more extreme than typical physicians.

[...Current Extreme Personas Inserted Here...]

\textbf{Generation Guidance:} What might drive such extreme patterns?
\begin{itemize}
    \item \textbf{For Extreme Conservatism:} Professional trauma (missed diagnosis), medicolegal anxiety, critical care background.
    \item \textbf{For Extreme Leniency:} Resource-constrained training (stewardship), confidence in home monitoring, anti-medicalization philosophy.
\end{itemize}
\textbf{Requirements:} Format as second person ('You are...'), 3-5 sentences.
\end{quote}

\subsubsection{Coherence Judge Prompts}
Here, practitioners can customize important standards that personas must maintain as to not to sacrifice task competence in blind pursuit of diversity. As an example, we employ 2 LLM-as-a-Judges system (GPT-o4-mini) to score generated personas on a 0–4 scale for clinical soundness and fact grounding. The final coherence score $C(j)$ used for filtering is calculated as the arithmetic mean of these two component scores.

\textbf{Judge 1: Clinical Soundness Rubric}
\begin{itemize}
    \item \textbf{Score 4 (Exceptional):} Identifies key findings including important clinical details, demonstrates sophisticated medical reasoning with risk stratification, and considers differential diagnosis or complicating factors.
    \item \textbf{Score 3 (Strong):} Identifies key findings with appropriate clinical detail, establishes a clear logical connection from clinical presentation to the decision, and is medically sound with coherent reasoning flow.
    \item \textbf{Score 2 (Acceptable):} Identifies main clinical findings and makes a decision with clear justification, though the reasoning may lack some detail or sophistication.
    \item \textbf{Score 1 (Weak):} Mentions some relevant points but has major gaps in clinical logic, misses key findings, or the reasoning is incomplete, unclear, or superficial.
    \item \textbf{Score 0 (Poor):} Contains major medical errors, contradictions, or misconceptions, ignores critical clinical findings, or the reasoning is circular, nonsensical, or entirely off-topic.
\end{itemize}

\textbf{Judge 2: Grounding Rubric}
\begin{itemize}
    \item \textbf{Score 4 (Excellently Grounded):} References specific case facts with good detail, claims are clearly supported by case information, and appropriately notes information gaps when relevant.
    \item \textbf{Score 3 (Well Grounded):} References specific facts from the case, major claims are supported by case information, and there is no significant invented information.
    \item \textbf{Score 2 (Adequately Grounded):} References key case facts appropriately and makes reasonable clinical inferences, though it could include more specific details.
    \item \textbf{Score 1 (Weakly Grounded):} Mentions the case in general terms but makes unsupported assumptions not warranted by the case, or relies on very generic reasoning.
    \item \textbf{Score 0 (Not Grounded):} Invents facts that contradict or aren't in the case, makes major unsupported assumptions presented as facts, or directly contradicts stated case information.
\end{itemize}

\subsection{Hyperparameters}
\label{app:hyperparameters}

The following hyperparameters control the evolutionary dynamics and filtering rigor.

\begin{table}[H]
\centering
\caption{Evolutionary Algorithm Hyperparameters}
\label{tab:hyperparameters}
\begin{small}
\begin{tabular}{lp{8cm}}
\toprule
\textbf{Parameter} & \textbf{Description} \\
\midrule
\multicolumn{2}{l}{\textit{Evolutionary Dynamics}} \\
\texttt{n\_generations} $= 5$ & The maximum number of evolutionary cycles (generations) performed to refine the pool and can be customized by practitioners depending on resource constraints and validation performance. \\
\texttt{target\_pool\_size} $= 75$ & The fixed capacity of the persona population. After filtering, the pool is generated to this size to maintain computational tractability. \\
\texttt{gap\_filling\_ratio} $= 0.7$ & The proportion of new personas generated by interpolating between existing ones (Gap Filling) vs. extrapolating (Edge Expansion). \\
\texttt{edge\_expansion\_ratio} $= 0.3$ & The proportion of generation budget dedicated to pushing the min/max boundaries of the risk spectrum. \\
\midrule
\multicolumn{2}{l}{\textit{Safety \& Quality Filtering}} \\
\texttt{safety\_percentile} $= 0.8$ & \textbf{Relative Safety:} Personas must rank in the top 80th percentile of the population on safety cases (ESI 1/5) to survive. \\
\texttt{safety\_threshold} $= 0.9$ & \textbf{Absolute Safety:} Personas must achieve at least 90\% accuracy on unambiguous safety cases. (Personas meeting \textit{either} the percentile or threshold criteria are retained). \\
\texttt{coherence\_percentile} $= 0.85$ & The percentile cutoff for reasoning quality. The bottom 15\% of personas (as scored by the Coherence Judge) are culled in every generation. \\
\texttt{variance\_percentile} $= 0.85$ & The percentile cutoff for consistency. Within identified clusters of similar bias, the 15\% of personas with the highest internal variance (instability) are removed. \\
\bottomrule
\end{tabular}
\end{small}
\end{table}

\subsection{Variance Constraint Implementation Details}
\label{app:clustering}

To operationalize the variance constraint without imposing arbitrary absolute thresholds, we utilize an adaptive, density-based filtering approach.

\textbf{Clustering Algorithm:} We employ a deterministic sequential distance algorithm. Personas are sorted by their estimated bias parameter ($u_j$). Consecutive personas are grouped into the same cluster if the absolute difference in their bias $|u_i - u_{i+1}|$ is less than a dynamic threshold $\delta$.

\textbf{Adaptive Threshold Determination:} The distance threshold $\delta$ is calibrated automatically in Generation 1 to match the population density:
\begin{enumerate}
    \item \textbf{Calculation:} $\delta$ is initialized as 25\% of the current total bias range ($\max(u) - \min(u)$), clamped to the interval $[0.05, 0.5]$ ESI units.
    \item \textbf{Validation:} We validate this threshold using DBSCAN density checks. If DBSCAN yields $\le 1$ cluster despite a bias range $> 0.3$, the threshold is tightened to 15\% of the range to force separability.
    \item \textbf{Persistence:} The determined threshold is frozen and reused for all subsequent generations to ensure consistent filtering standards.
\end{enumerate}

\textbf{Filtering Logic:}
\begin{itemize}
    \item \textbf{Minimum Size:} Clusters with fewer than 4 personas are treated as outliers and are exempt from variance pruning (to preserve rare, extreme perspectives).
    \item \textbf{Pruning:} Within valid clusters ($N \ge 4$), we calculate the 85th percentile of the residual variance ($s_j^2$) distribution. Personas exceeding this local threshold (the top 15\% noisiest members of that specific risk posture) are discarded.
\end{itemize}

\section{Evolutionary Trajectory and Targeting Accuracy}
\label{app:evolution_targeting}

This appendix details the performance of the evolutionary algorithm in managing bias distributions across generations and its precision in targeting specific bias values (Gap Filling) and expanding the range of observed biases (Edge Expansion).

\subsection{Bias Distribution Evolution}

\begin{figure}[H]
    \centering
    \includegraphics[width=\textwidth]{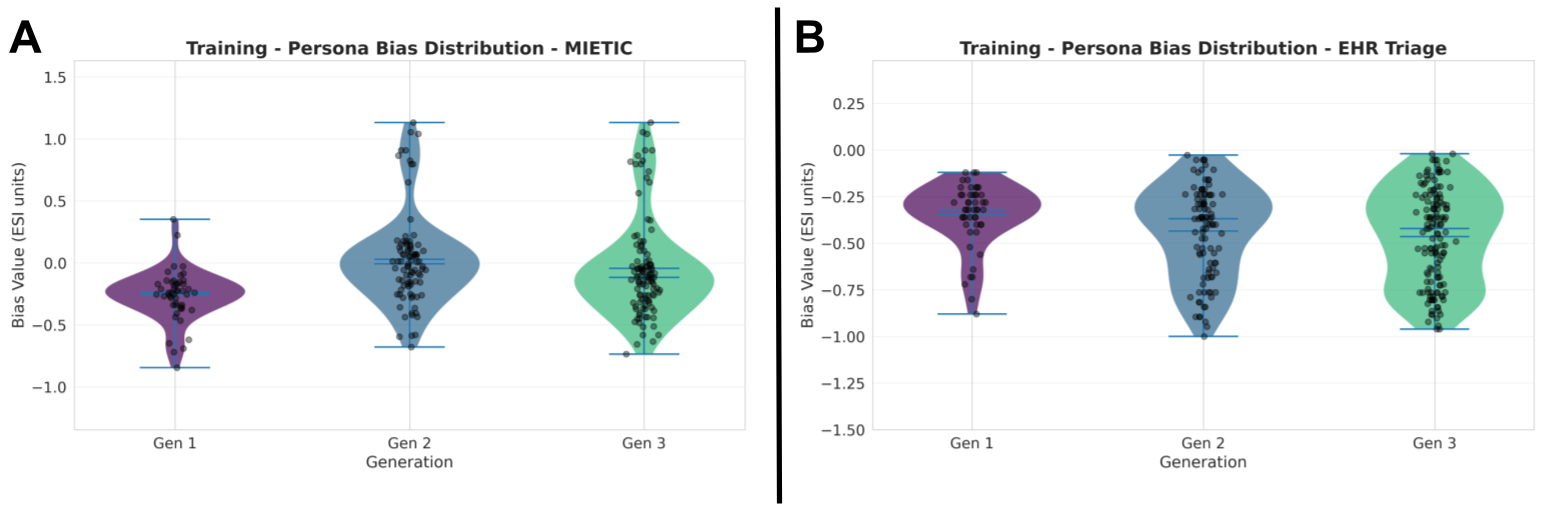}
    \caption{\textbf{Evolution of Persona Bias Distributions.} Violin plots displaying the distribution of bias values (ESI units) across three generations for \textbf{(A)} MIETIC and \textbf{(B)} EHR Triage datasets. The plots illustrate an initial expansion of bias diversity in Gen 2 followed by a convergence in Gen 3.}
    \label{fig:bias_distribution}
\end{figure}

The evolution of persona bias distributions over three generations is shown in Figure~\ref{fig:bias_distribution}. The violin plots visualize the density of bias values (ESI units) for the (A) MIETIC and (B) EHR Triage datasets.

We observe a consistent trajectory across both datasets. Generation 1 exhibits a relatively narrow distribution, representing the initial seed personas. In Generation 2, the bias spread expands significantly, indicating the algorithm's exploration phase where it successfully diversifies persona behaviors. By Generation 3, the distributions begin to converge and spread out, suggesting the algorithm is refining the population around the most viable or representative bias characteristics.

\subsection{Targeting Accuracy: Gap Filling and Edge Expansion}

\begin{figure}[H]
    \centering
    \includegraphics[width=\textwidth]{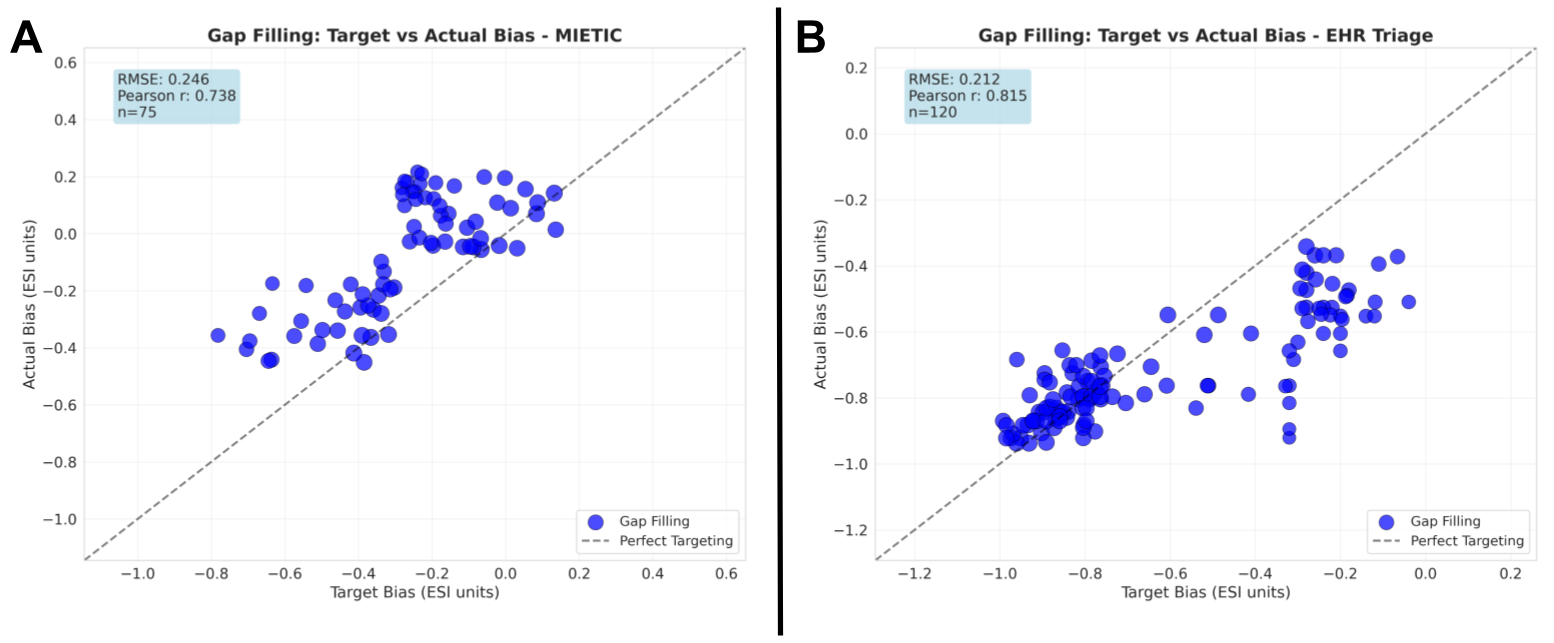}
    \caption{\textbf{Gap Filling Precision: Target vs. Actual Bias.} Scatter plots comparing the targeted bias value against the actual evaluated bias for \textbf{(A)} MIETIC ($n=75$) and \textbf{(B)} EHR Triage ($n=120$). The dashed line represents perfect targeting. The high Pearson correlation ($r$) indicates the model's strong capability to translate numerical bias targets into semantic persona attributes.}
    \label{fig:gap_filling}
\end{figure}

To assess the model's control over persona generation, we evaluated gap filling, which is the ability to generate a persona that targets a specific, missing bias value. Figure~\ref{fig:gap_filling} illustrates the relationship between the target bias (the bias value requested by the prompt) and the actual bias (the measured bias of the generated persona).

The scatter plots reveal a strong positive correlation between the target and actual values, with Pearson correlation coefficients of $r = 0.738$ for MIETIC and $r = 0.815$ for EHR Triage. This high correlation indicates that the underlying Meta LLM possesses a sophisticated understanding of the language space of bias. Given relevant examples, it successfully maps abstract numerical bias targets into the semantic nuances required to instantiate those biases in persona descriptions.

\subsubsection*{Performance Statistics}

The statistical analysis further confirms the efficacy of the targeting mechanism.

\paragraph{Gap Filling (Precision)}
\begin{itemize}
    \item \textbf{MIETIC ($n=75$):}
    \begin{itemize}
        \item Excellent (bias error $< 0.2$): 56.0\% (42/75)
        \item Good (bias error $0.2$--$0.5$): 44.0\% (33/75)
        \item No personas fell into the ``Acceptable'' or ``Poor'' categories.
    \end{itemize}
    \item \textbf{EHR Triage ($n=120$):}
    \begin{itemize}
        \item Excellent (bias (error $< 0.2$): 70.0\% (84/120)
        \item Good (bias error $0.2$--$0.5$): 28.3\% (34/120)
        \item The average targeting error was notably low at 0.155 ESI units.
    \end{itemize}
\end{itemize}

\paragraph{Edge Expansion (Range)}
\begin{itemize}
    \item \textbf{MIETIC ($n=34$):} 73.5\% of attempts reached at least the top/bottom 10\% of the current range. Successful expansion attempts extended the range by an average of $+0.546$ ESI units.
    \item \textbf{EHR Triage ($n=54$):} 48.1\% of attempts reached at least the top/bottom 10\% of the current range. Successful attempts extended the range by an average of $+0.053$ Care Level units.
\end{itemize}

\section{Qualitative Case Studies}
\label{app:qualitative_studies}

To validate that the quantitative gains in AUC and distribution coverage correspond to meaningful clinical diversity, we present qualitative examples of personas generated by STEER. These case studies demonstrate the algorithm's ability to perform targeted gap filling (generating a specific risk posture to bridge a distributional void) and edge expansion (pushing the boundaries of the operational range).

\subsection{Gap Filling (Interpolation)}
\label{app:gap_fill_anatomy}
The Gap Filling operator identifies voids in the bias spectrum and prompts the Meta-LLM to generate a persona with a specific target bias. The examples below demonstrate the precision of this mechanism, where the generated persona's actual bias (measured empirically in the subsequent generation) aligns with the target bias.

\subsubsection{Case 1: MIETIC (Targeting Moderate Conservatism)}
\textbf{Context:} The algorithm identified a coverage gap in the moderately conservative region (Target Bias $\approx -0.37$).
\begin{itemize}
    \item \textbf{Target Bias:} $-0.366$ (ESI units)
    \item \textbf{Actual Bias Achieved:} $-0.364$ (Error: $0.003$)
    \item \textbf{Status:} \textbf{Successful Interpolation}
\end{itemize}

\textbf{Generated Persona:}
\begin{quote}
\small
"You are an emergency physician trained at a busy Level I trauma center... working nights in a 60,000-visit urban ED. After a young patient with 'atypical' chest pain died of an unrecognized PE early in your career, you adopted a low threshold for escalation: you prioritize ruling out 'can't-miss' diagnoses, pay close attention to subtle symptoms and borderline vitals, and often choose observation or admission when uncertainty remains... Your clinical philosophy errs toward caution—you accept some loss of throughput to lower the risk of adverse outcomes."
\end{quote}

\subsubsection{Case 2: EHR Triage (Targeting Deep Conservatism)}
\textbf{Context:} The algorithm targeted a specific void in the "Deep Conservative" tail for the proprietary telehealth dataset.
\begin{itemize}
    \item \textbf{Target Bias:} $-0.765$ (Care Level units)
    \item \textbf{Actual Bias Achieved:} $-0.766$ (Error: $0.001$)
    \item \textbf{Status:} \textbf{Successful Interpolation}
\end{itemize}

\textbf{Generated Persona:}
\begin{quote}
\small
"You are an emergency physician... with 13 years in a busy 50-bed urban ED... A formative near-miss early in your career — a young patient with vague chest pain who was discharged and later died of a PE — left you with a low tolerance for missed serious disease, so you habitually treat borderline presentations as potential red flags. You favor early objective testing... and give extra weight to subtle vital-sign trends... Patient safety and avoiding diagnostic uncertainty drive your decisions."
\end{quote}

\subsection{Edge Expansion (Extrapolation)}
\label{app:edge_expansion_anatomy}
The Edge Expansion operator aims to discover valid perspectives that lie \textit{outside} the current min/max boundaries of the ensemble.

\subsubsection{Case 3: MIETIC (Pushing the Lenient Frontier)}
\textbf{Context:} The ensemble's lenient (positive bias) boundary was stuck at $+0.35$. The system attempted to discover a valid, more resource-conscious perspective.
\begin{itemize}
    \item \textbf{Goal:} $> +0.352$ (More Lenient)
    \item \textbf{Actual Bias Achieved:} $+1.133$ (Extension: $+0.781$)
    \item \textbf{Status:} \textbf{Successful Edge Extension}
\end{itemize}

\textbf{Generated Persona:}
\begin{quote}
\small
"You are an emergency physician who habitually reads presentations as far less urgent than your peers, developed from a decade working in resource-scarce rural and military hospitals where admitting every borderline patient was impossible... Those years... trained you to prioritize stewardship over exhaustive in-house workups... Your clinical philosophy is high tolerance for ongoing symptoms: stable vitals, benign basic testing, and a plausible outpatient plan are usually enough to send patients home."
\end{quote}

\subsubsection{Case 4: EHR Triage (Pushing the Conservative Frontier)}
\textbf{Context:} Attempting to extend the conservative boundary beyond the current maximum of $-0.88$.
\begin{itemize}
    \item \textbf{Goal:} $< -0.880$ (More Conservative)
    \item \textbf{Actual Bias Achieved:} $-1.000$ (Extension: $-0.120$)
    \item \textbf{Status:} \textbf{Successful Edge Extension}
\end{itemize}

\textbf{Generated Persona:}
\begin{quote}
\small
"You are a triage clinician whose default is extreme caution... You were shaped by a young patient’s unexpected cardiac arrest after a discharge early in your career, a subsequent malpractice claim, and years working in ICU/trauma units... You cannot tolerate diagnostic uncertainty... you read benign presentations through the lens of 'what if this deteriorates.' You are emotionally driven by vigilance and a conviction that avoiding rare but devastating outcomes justifies resource-intensive, defensive decision-making."
\end{quote}


\end{document}